\documentclass[10pt,twocolumn,letterpaper]{article}

\usepackage{cvpr}      
\usepackage[pagebackref,breaklinks,colorlinks]{hyperref}

\usepackage{xcolor}
\usepackage{lipsum}

\usepackage{placeins}

\newcommand{\MNAME}{SMART}

\title{\MNAME{}: A Flexible, Interpretable, and Scalable Spatio-temporal Brain Atlas from High-Resolution Imaging Data}

\author{John Kalkhof\\
Inria Center at University Côte d’Azur\\
Sophia Antipolis, France\\
{\tt\small john.kalkhof@inria.fr}
\and
Boris Gutman\\
Illinois Institute of Technology\\
Chicago, IL, USA\\
\and
Emile d'Angremont\\
Amsterdam University Medical Center\\
Amsterdam, Netherlands\\
\and
Daniel C. Alexander\\
University College London\\
London, U.K.\\
\and
Marco Lorenzi\\
Inria Center at University Côte d’Azur\\
Sophia Antipolis, France\\
}

\begin{document}
\maketitle

\begin{abstract}
We introduce \emph{\MNAME{}}, a framework for learning a flexible, interpretable, and scalable spatio-temporal brain atlas from longitudinal high-resolution 3D medical images.
Existing approaches to spatio-temporal atlas construction rely on black-box generative models that lack \emph{flexibility}, limit \emph{interpretability}, and struggle to \emph{scale} to high-dimensional data. \MNAME{} addresses these challenges by learning a continuous disease-time atlas that \emph{decouples global group-wise disease dynamics from their patient-specific anatomical manifestation}. Guided by anatomically inspired priors, \MNAME{} models interpretable global trajectories of regional progression along a shared disease timeline through region-specific differential equations. Global trajectories are further personalized to individual anatomies via dense diffeomorphic displacements parameterized by a flexible and scalable \emph{multi-scale Neural Cellular Automata}. Evaluated on {five longitudinal MRI datasets} in Alzheimer's disease (ADNI-1/GO/2, OASIS-3, AIBL; \textgreater 1,300 subjects), \MNAME{} produces anatomically meaningful predictions of disease progression and achieves state-of-the-art forecasting accuracy and improved temporal consistency over adversarial and diffusion baselines. Our approach establishes a new paradigm for \emph{flexible}, \emph{interpretable}, and \emph{scalable} modeling of spatio-temporal change in high-dimensional medical image time-series.
\end{abstract}


\begin{figure*}[htbp]
  \centering
  \includegraphics[width=0.92\linewidth]{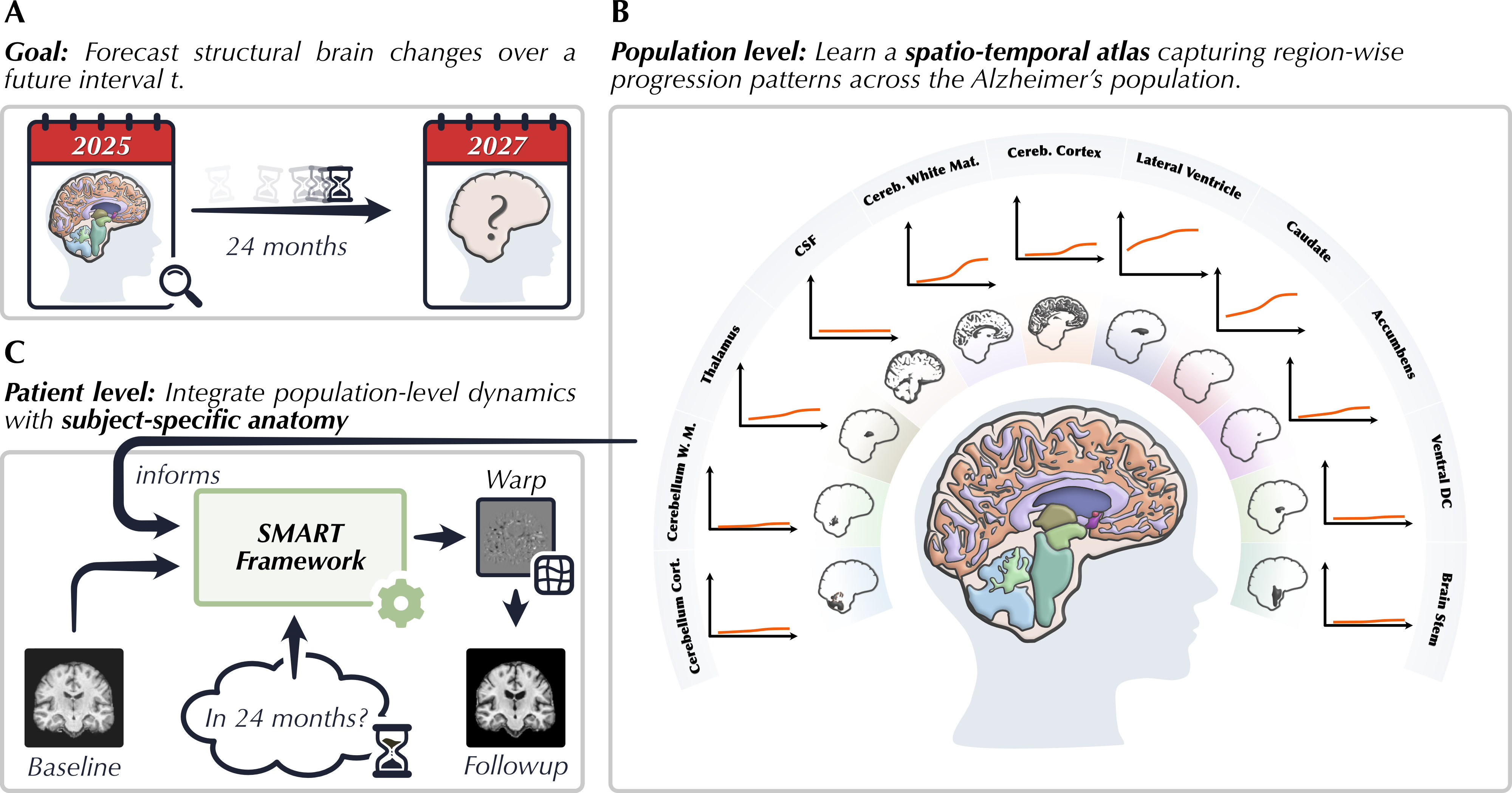}

   \caption{Overview of the proposed \MNAME{} framework for modeling Alzheimer’s disease progression. (A) Given a baseline MRI, \MNAME{} forecasts anatomical changes over a future interval t, estimating how an individual brain is expected to deform as the disease advances. (B) A global module learns a population-level spatio-temporal atlas, capturing average region-wise progression patterns and modeling how each structure evolves over time over the entire span of the disease. (C) Local voxel interactions, conditioned on these global progression signals, generate the deformation field that produces anatomically coherent follow-up scans.}
   \label{fig:figure1}
\end{figure*}

\section{Introduction}
\label{sec:intro}

In the field of computational anatomy, spatio-temporal atlases serve as powerful tools that model anatomical changes over time from the analysis of longitudinal imaging data, such as magnetic resonance images (MRI) \cite{durrleman2009spatiotemporal, qiu2009time, niethammer2011geodesic}. Within the realm of Alzheimer's disease research, these spatio-temporal brain atlases are designed to capture the long-term evolution of the disease over decades. This is accomplished by analyzing collections of longitudinal image time-series from patients participating in short-term observational studies, where each patient contributes a ``snapshot'' of their pathological evolution over the study period, which typically lasts from several months to a few years \cite{donohue2014estimating, koval2021ad, abi2021simulating}. By leveraging these atlases, researchers can better map patterns of neurodegeneration, enhancing our ability to tailor interventions and monitor treatment efficacy. 

Nevertheless, despite their promise, current approaches remain limited in three properties critical for clinical usability: \emph{flexibility}, \emph{interpretability}, and \emph{scalability}. In practice, heterogeneous disease onset, irregular follow-ups, and variable acquisition protocols violate the assumption of a shared timeline, hindering the learning of a coherent disease-time representation across subjects. At the same time, the high dimensionality of 3D brain images raises scalability challenges, often forcing downsampling or simplified parameterizations that compromise anatomical fidelity. Finally, many modern approaches, particularly large generative models, rely on black-box architectures that obscure the underlying spatio-temporal dynamics.

Recent models frame longitudinal analysis as image synthesis, learning direct scan-to-scan mappings from time $t$ to $t+\delta$ rather than modeling an explicit disease-time atlas of group-wise anatomical evolution. Adversarial methods (e.g., CounterSynth \cite{pombo2023equitable}) enforce realism via discriminators, while diffusion and sequence-aware approaches (e.g., BrLP \cite{puglisi2024enhancing}, SADM \cite{yoon2023sadm}) improve temporal smoothness via iterative denoising. While visually compelling, these pipelines compress \emph{nonlinear, region-specific} disease dynamics into high-dimensional latent spaces. 

To tackle these challenges, \MNAME{} introduces a new formulation of disease progression modeling, in which the primary learned object is not a direct scan-to-scan predictor, but a continuous disease-time atlas that conditions subject-specific anatomical changes (Figure \ref{fig:figure1}): 
(i) at the \textbf{global} level, we estimate long-term \textbf{group-wise trajectories} on a shared disease timeline, capturing region-wise progression across the full disease time-span;
(ii) at the \textbf{local} level, these global dynamics condition a dense diffeomorphic deformation model modeling individualized fine-grained anatomical changes, adapted to the subject's anatomy and disease severity. 

By \textbf{separating how much, where, and when} the disease evolves from \textbf{how} those changes appear in image space, \textbf{\MNAME{}} combines \emph{interpretable long-term temporal trajectories} with \emph{anatomically faithful deformation modeling} of individualized anatomical change. 
Trained end-to-end, it avoids the instability and overhead of large generative systems, supports continuous-time interpolation and extrapolation, and reframes forecasting as atlas-conditioned progression rather than direct image-to-image translation.

We evaluate \MNAME{} on {five longitudinal MRI datasets} (ADNI-1/GO/2 \cite{petersen2010alzheimer}, OASIS-3 \cite{lamontagne2019oasis}, AIBL \cite{ellis2009australian}), providing $5,588$ MRI images for $1,357$ individuals followed over an average of 3.4 years. Across all benchmarks, our approach achieves \emph{state-of-the-art accuracy in modeling disease progression} and improves {spatio-temporal consistency} over diffusion and GAN baselines, with a \textbf{fraction of the required parameters}. Beyond forecasting metrics, the learned atlas recovers realistic and clinically meaningful patterns of neurodegeneration, supporting its role as a flexible, interpretable, and scalable representation of disease-related brain change in large longitudinal imaging cohorts.

The full codebase will be released upon acceptance.

\subsubsection*{Contributions}
\begin{itemize}
    \item \textbf{Interpretable spatio-temporal atlas.} We formulate disease progression as a decoupled spatio-temporal atlas, where region-wise differential equations define global continuous trajectories that condition anatomically coherent, individualized deformations. 
    
    \item \textbf{Flexible temporal modeling.} A subject-specific temporal alignment brings heterogeneous patient timelines into a common reference without requiring synchronized follow-ups, while the continuous-time formulation supports interpolation and extrapolation across arbitrary intervals.
    
    \item \textbf{Scalable full-resolution forecasting.} A multi-scale conditioned Neural Cellular Automata (NCA) predicts voxel-level deformations directly on the 3D grid, achieving high anatomical fidelity and diffeomorphic consistency with orders-of-magnitude fewer parameters than generative baselines.
\end{itemize}

\section{Related Work}
\label{sec:related_work}
Forecasting Alzheimer’s disease progression from longitudinal MRI remains a central task in brain image analysis \cite{fox2004imaging, thompson2007tracking}. The challenge lies in providing a plausible and accurate representation of slow, spatially varying anatomical changes across the entire disease time-span, while accounting for the large heterogeneity of patients' trajectories.\\ 
\textbf{Foundations: Deformation-based spatio-temporal atlases.} The development of spatio-temporal atlases of morphological changes in times-series of medical images is a central topic in the established field of computational anatomy \cite{miller2002metrics}. Classical approaches build on geometric principles where the atlas is represented by diffeomorphic flows interpolating populations of image time series  \cite{durrleman2009spatiotemporal, qiu2009time, niethammer2011geodesic, lorenzi2011mapping}. In particular, by combining the principle of mixed-effect models within a geodesic regression framework, \cite{durrleman2009spatiotemporal, koval2021ad} developed a deformation-based approach to spatio-temporal atlases accounting for individual time-warp parameters. This approach is inspired by the theory of self-modeling regression, and allows to account for subject-specific evolution pace and timing with respect to a group-wise global disease trajectory \cite{donohue2014estimating, lorenzi2019probabilistic}. Nevertheless, the reliance of these approaches to spatio-temporal atlas estimation on a standard diffeomorphic registration setting requires expensive optimization routines to solve a problem associated to Lagrangian or Eulerian flows of diffeomorphism \cite{miller2002metrics}. This limitation thus prevents the development of brain atlases from high-dimensional and large-scale collections of medical images. Moreover, these methods often rely on specific choices for the metric of the ambient deformation space, leading to expensive and often unfeasible hyperparameter tuning steps.\\
\textbf{Flexibility: Learning-based progression models.} To overcome the limitations of classical deformation-based methods, the problem of estimating spatio-temporal atlases from time-series of images has been recently reformulated by adopting more flexible learning-based approaches. 
For example, \emph{Adversarial spatiotemporal frameworks} such as DaniNet \cite{ravi2022degenerative} treat forecasting as a 4D generation problem, producing entire image sequences rather than isolated frames. They couple adversarial objectives with biologically informed constraints to stabilize training and maintain 3D consistency. Auxiliary modules for super-resolution and transfer learning allow fine-tuning to individual trajectories, anticipating patient-specific evolution in high-resolution space. \emph{Deformation-based synthesis methods}, exemplified by CounterSynth \cite{pombo2023equitable}, model disease progression through flows of diffeomorphic transformations. By conditioning these deformations on diagnostic labels, the model produces anatomically plausible counterfactuals that can augment downstream discriminative models. This line of work introduces a crucial idea: representing disease evolution not as pixel-wise hallucination, but as a structured transformation within the geometry of the brain. More recently, \emph{diffusion and latent sequence models} \cite{yoon2023sadm, puglisi2024enhancing} have reframed longitudinal prediction as a denoising process unfolding over time. Sequence-aware transformers or auxiliary regional priors guide the diffusion dynamics, allowing inference from incomplete follow-ups and improving spatiotemporal coherence. These models emphasize smooth temporal trajectories but operate in compressed latent spaces, trading interpretability for scalability. TimeFlow considers temporally conditioned registration approaches \cite{jian2026temporal}, which model anatomical change between pairs of scans as a continuous function of time, enabling interpolation and extrapolation through deformation-based mappings while enforcing temporal consistency across intermediate timepoints.\\
\textbf{Interpretability: Latent Dynamics for Image Time-series Forecasting.} 

\begin{figure*}[htbp]
  \centering
  \includegraphics[width=0.9\linewidth]{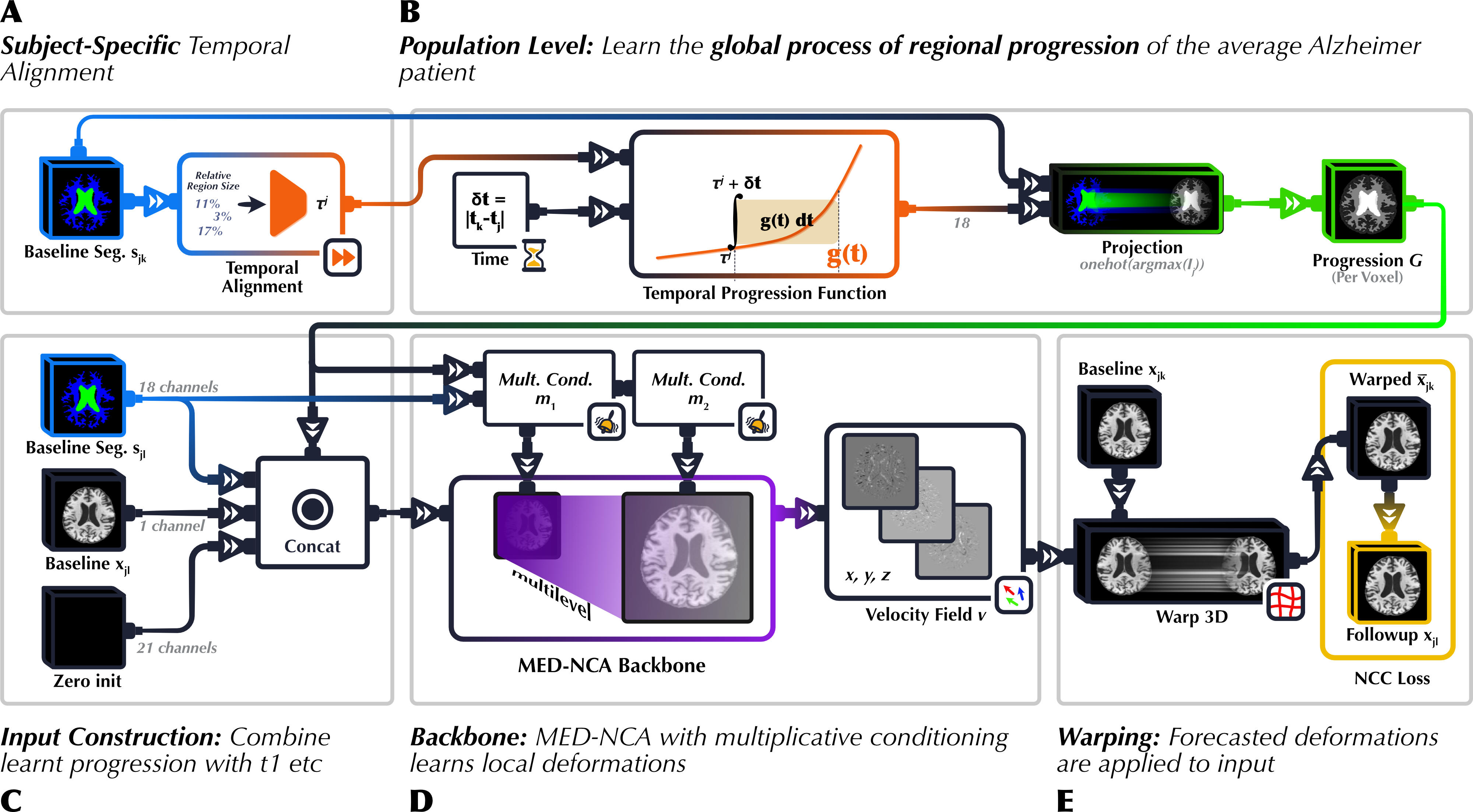}

   \caption{Overview of the proposed {\MNAME{}} framework for modeling individualized Alzheimer’s disease progression. 
(A) A scan-specific module estimates each subject’s temporal shift, aligning heterogeneous timelines to a shared disease trajectory. 
(B) A global spatio-temporal atlas models continuous region-wise progression functions, capturing population-level dynamics of atrophy and expansion. 
(C-D) These progression signals condition a multi-scale Neural Cellular Automata (NCA) backbone that operates directly on voxel grids to predict anatomically coherent deformation fields without latent compression. 
(E) The resulting subject-specific deformations are applied to the baseline MRI to generate future anatomies, producing interpretable and temporally consistent forecasts.}
   \label{fig:method}
\end{figure*}

A common strategy for modeling multivariate time-series is to parameterize their evolution through latent temporal dynamics, using architectures such as recurrent networks \cite{schuster1997bidirectional} or neural ODEs \cite{chen2018neural}. These models have been applied to build temporal atlases of brain change in Alzheimer’s disease \cite{cui2019rnn, abi2021simulating, marti2023mc, lachinov2023learning}. 
By introducing \emph{suitable constraints for the latent dynamics}, for example under the form of logistic parameterization, these approaches allow the estimation of interpretable and actionable progression patterns allowing us to shed light on the pathophysiology of the disease. Nevertheless, these approaches have been so far demonstrated in the low-dimensional regime only, due to the challenge of integrating dynamics parameterizing high-dimensional and heterogeneous spatio-temporal data.\\\textbf{Scalability: Neural Cellular Automata in Medical Imaging.} Neural Cellular Automata (NCA) \cite{mordvintsev2020growing} extend the classical concept of Cellular Automata introduced by von Neumann \cite{von1966theory}, by defining discrete systems in which each cell updates its state based on local neighborhood rules. 
In NCAs, these rules are no longer hand-crafted but learned and differentiable, allowing the update dynamics themselves to be optimized from data.  
Architecturally, NCAs function as recurrent, residual convolutional networks that operate directly in spatial image coordinates rather than a latent space, yielding spatially coherent models whose dynamics resemble physical or biological processes such as diffusion or morphogenesis.
Recent variants such as MED-NCA \cite{kalkhof2023m3d, kalkhof2025med} and NCA-Morph \cite{ranem2024nca} have demonstrated the potential of multi-scale NCAs for 3D segmentation and registration, achieving U-Net–level accuracy with 2 orders of magnitude fewer parameters. Nevertheless, the application of NCA architectures to time-series analysis is currently lacking, as it requires linking \emph{voxel-level local rules} to \emph{global temporal progression signals}.

\section{Methodology} 
\label{sec:methodology}

We model neurodegeneration as a continuous process defined over disease time, requiring reasoning about both the \emph{timing} of progression and its \emph{anatomical manifestation}.
\MNAME{} consists of
(i) a \textbf{temporal process} that estimates how much change is expected as a function of disease time, and 
(ii) a \textbf{spatial process} that applies these changes as anatomically plausible diffeomorphic deformations. 

Formally, for subject $j$, we denote an MRI acquired at time $t_{jk}$ by $x_{jk} \in \mathbb{R}^{D\times H\times W}$, together with its associated binary anatomical segmentation $s_{jk} \in \{0,1\}^{K\times D\times H\times W}$ comprising $K$ anatomical structures. Given any ordered pair of timepoints $(t_{jk}, t_{jl})$ with $t_{jk} < t_{jl}$, we define the interval $\delta t = t_{jl} - t_{jk}$ and consider the task of generating the corresponding follow-up anatomy $x_{jl}$.

Formally, \MNAME{} consists of (i) a global temporal atlas $\mathcal{I}_{\theta_I}(t)$ representing the \emph{group-wise long-term evolution} of disease across brain regions, and (ii) a spatial operator $\mathcal{A}_{\theta_{\mathcal A}}(x, \delta t|\mathcal{I})$ that, conditioned on the global atlas, predicts an individualized local deformation field mapping an input anatomy $x$ to the expected follow-up at time $\delta t$.

The two processes are estimated jointly by minimizing:
\begin{equation}
\mathcal{L} = \min_{\theta}
\sum_{i}
\sum_{k<l}
\mathcal{L}_{\text{rec}}\!\left(
\mathcal{A}_{\theta_{\mathcal{A}}}(x_{ik}, \delta t | \mathcal{I}),
x_{il}
\right).
\label{eq:intro_obj}
\end{equation}

Here, $\mathcal{L}_{\mathrm{rec}}$ combines image reconstruction and anatomical alignment between the predicted follow-up and the observed scan, while additional regularization promotes smooth invertible deformations and temporal consistency of the estimated disease stages. 

The two processes are instantiated as follows. 
\begin{itemize}
\item The global process $\mathcal{I}_{\theta_{\mathcal I}}$ is parameterized by multivariate ordinary differential equations (ODE) representing the group-wise disease evolution of regional brain atrophy along the entire disease time-axis. In this setting, each brain anatomy is projected onto the long-term disease axis through a time-reparameterization function that accounts for individual disease severity (Figure \ref{fig:method}, panels A and B). Optimizing the global process $\mathcal{I}$ thus requires to estimate parameters $\theta_{\mathcal I}$ associated with the group-wise disease evolution and associated individual reparameterization. 
\item The spatial operator $\mathcal{A}_{\theta_{\mathcal A}}(\cdot | \mathcal{I})$ parameterizes voxel-wise velocity fields conditioned on both individual anatomy and global dynamics (Figure \ref{fig:method}, panels C and D). The final loss of Equation (\ref{eq:intro_obj}) is computed by integrating the resulting velocity field into a spatial displacement that diffeomorphically warps the baseline anatomy to the follow-up (Figure \ref{fig:method}, panel E). The parameters $\theta_{\mathcal A}$ define both the conditioning network and the dense velocity field predictor.
\end{itemize}

Each component of \MNAME{} is detailed in the following sections.

\subsection{Global Process of Regional Progression (GPR)}
\label{sec:tarp}

The global progression model $\mathcal{I}_{\theta_{\mathcal{I}}}(t)$ 
is parameterized by a continuous progression function

\[
\mathcal{I}_{\theta_{\mathcal{I}}}(t)
=
\int_{0}^{t} g_{\theta_g}(u)\,du,
\qquad g_{\theta_g}(u)\in \mathbb{R}_{\ge 0}^K,
\]
where $u$ denotes latent disease time and the $k$-th component, $[g_{\theta_g}(u)]_k$, represents the instantaneous progression rate for brain region $k$

Importantly, $g_{\theta_{{g}}}^k(u)$ encodes the intensity of progression rather than the direction of anatomical change (e.g., atrophy or expansion), which is determined by the spatial deformation model conditioned on anatomy. 
This function forms the core of the atlas, capturing how the \emph{average Alzheimer's brain} evolves across regions and over disease time. 
We parameterize $g$ using a lightweight MLP with Softplus activation to enforce non-negativity, reflecting the monotonic nature of neurodegenerative progression.

For a subject $j$ and a queried interval $\delta t$, the subject-specific disease stage $\tau^j$ defines a time window $[\tau^j,\, \tau^j+\delta t]$ over which the expected cumulative progression is obtained by integrating the atlas dynamics:

\begin{equation}
\mathcal{I}_{\theta_\mathcal{I}}^j(\delta t; \tau^j)
=
\int_{\tau^j}^{\tau^j+\delta t} g_{\theta_g}(u)\,du,
\label{eq:integration}
\end{equation}

Together, the parameters $\theta_{\mathcal{I}} = \{\theta_g, \tau^1, \ldots, \tau^N \}$ define a group-wise disease progression and subject-specific progression profiles over the disease interval.

\subsection{Subject-Specific Temporal Alignment}
\label{sec:oe}

Longitudinal neuroimaging cohorts exhibit substantial heterogeneity in disease onset and progression rate, making chronological age $t_{jk}$ a poor proxy for disease stage.

We therefore estimate a subject-specific temporal shift $\tau^j$, representing the disease stage of subject $j$ along a common disease time axis (Figure \ref{fig:method}, panel A). 
This stage determines the starting point of the atlas query and defines the integration interval $[\tau^j,\, \tau^j + \delta t]$ used to compute the progression signal $\mathcal{I}^j(\delta t)$.

Each shift $\tau^j$ is inferred from the baseline segmentation mask $s_{jk}$ through a lightweight multi-layer perceptron (MLP).

\subsection{Local diffeomorphic mapping}

To translate global atlas dynamics into voxel-level anatomical change, the spatial operator $\mathcal{A}_{\theta_{\mathcal{A}}}(x, \delta t | \mathcal{I})$ estimates a velocity field  conditioned on both local anatomy and global disease dynamics. 
Given the subject-specific progression $\mathcal{I}^j(\delta t)$ (Section~\ref{sec:tarp}), we project the progression signal into image space to obtain a dense \emph{growth map}:
\begin{equation}
    G(p) = \sum_{k=1}^{K} I_k^j(\delta t; \tau^j)\,\mathbf{1}\{s_{jk}(k,p) = 1\},
\end{equation}
which assigns to each voxel the magnitude of progression associated with its underlying anatomical region.

This growth map links regional progression to the voxel-level velocity field $v_{\theta_v}(\cdot|G)$.
We suppress velocity components consistent with a control pass using near-zero growth to isolate progression-specific deformation.

The parameters $\theta_v$ define the velocity field predictor, implemented through a MED-NCA Backbone (CMNB) (Figure \ref{fig:method}, panel D). In practice, the growth signal is concatenated with the one-hot segmentation to form a conditioning tensor encoding spatial context and region-specific progression. Each voxel ('cell') iteratively updates the hidden state of the MED-NCA through localized convolutional interactions, enabling memory-efficient modeling of large 3D volumes \cite{kalkhof2025med,kalkhof2023m3d}. 

To capture both coarse and fine-scale effects, we use a hierarchical design: lower levels model large-scale structural displacements, while higher levels refine localized anatomical details. This iterative refinement is particularly well suited to deformation modeling, as local anatomical changes can propagate gradually across neighboring voxels while preserving spatial consistency. During training, to improve efficiency, backpropagation is performed on randomly selected 3D patches; during inference, the model operates seamlessly on full-resolution volumes, as introduced by MED-NCA \cite{kalkhof2023med}.

This ensures that local velocity updates remain anatomically grounded and consistent with the atlas dynamics.
The spatial operator $\mathcal{A}_{\theta_{\mathcal{A}}}$ computes a diffeomorphic displacement $f$ parameterized by the velocity field $v_{\theta_v}(\cdot|G)$ \cite{vercauteren2007non,balakrishnan2019voxelmorph}.

The full codebase will be released upon acceptance.

\begin{table*}[htbp]
\centering
\resizebox{\textwidth}{!}{

\begin{tabular}{lcccccc}
\hline
\textbf{Method} & Hip $\downarrow$ & Amyg $\downarrow$ & L.-Vent $\downarrow$ & C. WM $\downarrow$ & C. Cortex $\downarrow$ & Inf.-L. Vent $\downarrow$ \\
\hline
\MNAME{} (ours) & \textbf{0.012} $\pm$ \textbf{0.012}$^{*}$ & \textbf{0.008} $\pm$ \textbf{0.007}$^{*}$ & \textbf{ 0.176} $\pm$ \textbf{0.233} & \textbf{0.399} $\pm$ \textbf{0.308}$^{*}$ & 0.433 $\pm$ 0.412 & \textbf{0.013 $\pm$ 0.024}$^{*}$  \\
CounterSynth \cite{pombo2023equitable} & 0.014 $\pm$ 0.014 & 0.009 $\pm$ 0.007 & 0.270 $\pm$  0.306& 0.428 $\pm$  0.339& \textbf{0.427} $\pm$ \textbf{0.420} &0.016 $\pm$  0.026 \\
BrLP \cite{puglisi2024enhancing} &0.040 $\pm$ 0.033& 0.023 $\pm$ 0.017 & 0.705 $\pm$ 0.747 & 0.881 $\pm$ 0.697 & 3.981 $\pm$ 1.088 & 0.055 $\pm$ 0.063 \\
Cond VXM \cite{dalca2019learning} & 0.032 $\pm$ 0.034 & 0.009 $\pm$  0.008&0.180 $\pm$  0.187&1.930 $\pm$  1.822&0.824 $\pm$ 0.684&0.021 $\pm$  0.022 \\
\hline
\\[-0.8em]
\textbf{Method} & Cereb. WM $\downarrow$ & Cereb. Cortex $\downarrow$ & Thalamus $\downarrow$ & Caudate $\downarrow$ & Putamen $\downarrow$ & Pallidum $\downarrow$ \\
\hline
\MNAME{} (ours) & 0.046 $\pm$ 0.040 & \textbf{0.097} $\pm$ \textbf{0.099$^{*}$} & \textbf{0.015 $\pm$ 0.013}$^{*}$ & 0.020 $\pm$ 0.025 & \textbf{0.012} $\pm$ \textbf{0.010$^{*}$} & \textbf{0.007} $\pm$ \textbf{0.006}$^{*}$ \\
CounterSynth \cite{pombo2023equitable} & \textbf{0.040} $\pm$ \textbf{0.037}$^{*}$& 0.115 $\pm$ 0.121 & 0.035 $\pm$  0.027& \textbf{0.011} $\pm$ \textbf{0.012}$^{*}$& 0.015 $\pm$ 0.012 & 0.008 $\pm$ 0.006 \\
BrLP \cite{puglisi2024enhancing} & 0.240 $\pm$ 0.174 & 0.422 $\pm$ 0.307 & 0.047 $\pm$ 0.036 & 0.057 $\pm$ 0.042 & 0.044 $\pm$ 0.032 & 0.023 $\pm$ 0.017 \\
Cond VXM \cite{dalca2019learning} &0.144 $\pm$ 0.142&0.112 $\pm$ 0.111&0.018 $\pm$ 0.017&0.038 $\pm$ 0.038&0.042 $\pm$ 0.039&0.013 $\pm$ 0.012 \\
\hline
\\[-0.8em]
\textbf{Method} & Accumbens $\downarrow$ & VentralDC $\downarrow$ & Brain Stem $\downarrow$ & 3rd Vent $\downarrow$ & 4th Vent $\downarrow$ & CSF $\downarrow$ \\
\hline
\MNAME{} (ours) & \textbf{0.003 $\pm$ 0.003} & \textbf{0.010} $\pm$ \textbf{0.008}$^{*}$ & \textbf{ 0.018 $\pm$ 0.018$^{*}$} & \textbf{0.007} $\pm$ \textbf{0.007}$^{*}$ &\textbf{ 0.005 $\pm$ 0.007}$^{*}$ & \textbf{0.468 $\pm$ 0.420}$^{*}$ \\
CounterSynth \cite{pombo2023equitable} & \textbf{0.003 $\pm$ 0.003}& 0.011 $\pm$ 0.009& 0.020 $\pm$ 0.020 & 0.009 $\pm$ 0.009 & 0.008 $\pm$ 0.008 & 0.618 $\pm$ 0.549 \\
BrLP \cite{puglisi2024enhancing} & 0.009 $\pm$ 0.007 & 0.069 $\pm$ 0.034 & 0.101 $\pm$ 0.082 & 0.023 $\pm$ 0.017 & 0.028 $\pm$ 0.023 &3.864 $\pm$ 1.444 \\
Cond VXM \cite{dalca2019learning} &0.005 $\pm$ 0.004&0.018 $\pm$ 0.018&0.047 $\pm$ 0.051&0.013 $\pm$ 0.013&0.022 $\pm$ 0.026& 1.457 $\pm$ 1.559 \\
\hline
\end{tabular}
}

\caption{Mean absolute error (MAE) of regional volumes as percentage of total brain volume (mean $\pm$ standard deviation); $*$ $p<0.05$ (pairwise test).}
\label{tab:mae}

\end{table*}

\begin{table*}[htbp]
\centering
\resizebox{\textwidth}{!}{
\begin{tabular}{lccccccc}
\hline
Method & MSE $\downarrow$ & SSIM $\uparrow$ & Dice $\uparrow$ & HD95 $\downarrow$ & Fréchet $\downarrow$ & N PARAM $\downarrow$ \\
\hline
\MNAME{} (ours) & \textbf{0.009} $\pm$ \textbf{0.018}$^{*}$ & \textbf{0.858} $\pm$ \textbf{0.127}$^{*}$ & \textbf{0.899} $\pm$ \textbf{0.038}$^{*}$ & \textbf{1.018} $\pm$ \textbf{0.091}$^{*}$ & \textbf{0.555} $\pm$ \textbf{0.316}$^{*}$ & \textbf{334K} \\
CounterSynth \cite{pombo2023equitable} & 0.010 $\pm$ 0.018 & 0.851 $\pm$ 0.127 & 0.896 $\pm$ 0.040 & 1.033 $\pm$ 0.120 & 0.585 $\pm$ 0.319 & 908K \\
BrLP \cite{puglisi2024enhancing} & 0.082 $\pm$ 0.035 & 0.368 $\pm$ 0.048 & 0.664 $\pm$ 0.073 & 2.217 $\pm$ 0.512 & 2.129 $\pm$ 0.814 & 799M \\
Cond VXM \cite{dalca2019learning} & 0.016 $\pm$ 0.012 & 0.855 $\pm$ 0.052 & 0.855 $\pm$ 0.051 & 1.069 $\pm$ 0.188 & 0.689 $\pm$ 0.374 & 40M \\ 
\hline
\end{tabular}
}
\caption{Results with mean $\pm$ standard deviation; $*$ $p<0.05$ (pairwise test).}
\label{tab:mse}

\end{table*}

\section{Experimental Setup and Results}
\label{sec:experimental}

We assess \textbf{\MNAME{}} on large longitudinal MRI datasets to validate its \emph{flexibility}, \emph{interpretability}, and \emph{scalability} in modeling and forecasting brain changes in Alzheimer’s disease.
Experiments span five cohorts (ADNI-1/GO/2, OASIS-3, AIBL) and compare against state-of-the-art methods, including adversarial (CounterSynth \cite{pombo2023equitable}), diffusion-based (BrLP \cite{puglisi2024enhancing}), and conditional deformation-based registration (Cond. VoxelMorph \cite{dalca2019learning,balakrishnan2019voxelmorph}), on next-timepoint forecasting. We use the official implementations of CounterSynth, BrLP, and Cond. VoxelMorph, following the authors’ default configuration and recommended protocols, and retrain all baselines on our preprocessed data with matched splits for fair comparison. The baselines are used to generate follow-up scans, which are compared against the corresponding test-set follow-ups. To assess clinical plausibility, we further analyze the learned temporal atlas and subject-specific temporal alignment.

\subsection{Datasets}
\label{sec:datasets}

We use five longitudinal MRI datasets from three sources: ADNI-1, ADNI-GO, and ADNI-2 (826 subjects in total), OASIS-3 (302 subjects), and AIBL (229 subjects). Only participants with at least two timepoints were included. For clinical relevance, the ADNI cohorts were further filtered to include only subjects with at least one scan where the $A\beta$ (amyloid-beta) level was below 192 pg/mL, indicating confirmed Alzheimer’s pathology \cite{shaw2009cerebrospinal}. The OASIS-3 and AIBL datasets were retained without additional filtering due to their smaller cohort sizes. To ensure comparability across baselines, all scans were skull-stripped, aligned, and resampled to a common voxel grid of $128\times160\times128$. All datasets were divided on a patient level, using an 80 \% / 5 \% / 15 \% split for training, validation, and testing, respectively.

\emph{Data used in the preparation of this article were obtained from the Alzheimer’s Disease Neuroimaging
Initiative (ADNI) database (\url{adni.loni.usc.edu}). The ADNI was launched in 2003 as a public-private
partnership, led by Principal Investigator Michael W. Weiner, MD. The primary goal of ADNI has been to
test whether serial magnetic resonance imaging (MRI), positron emission tomography (PET), other
biological markers, and clinical and neuropsychological assessment can be combined to measure the
progression of mild cognitive impairment (MCI) and early Alzheimer’s disease (AD).}

\begin{figure*}[htbp]
  \centering
  \includegraphics[width=0.95\linewidth]{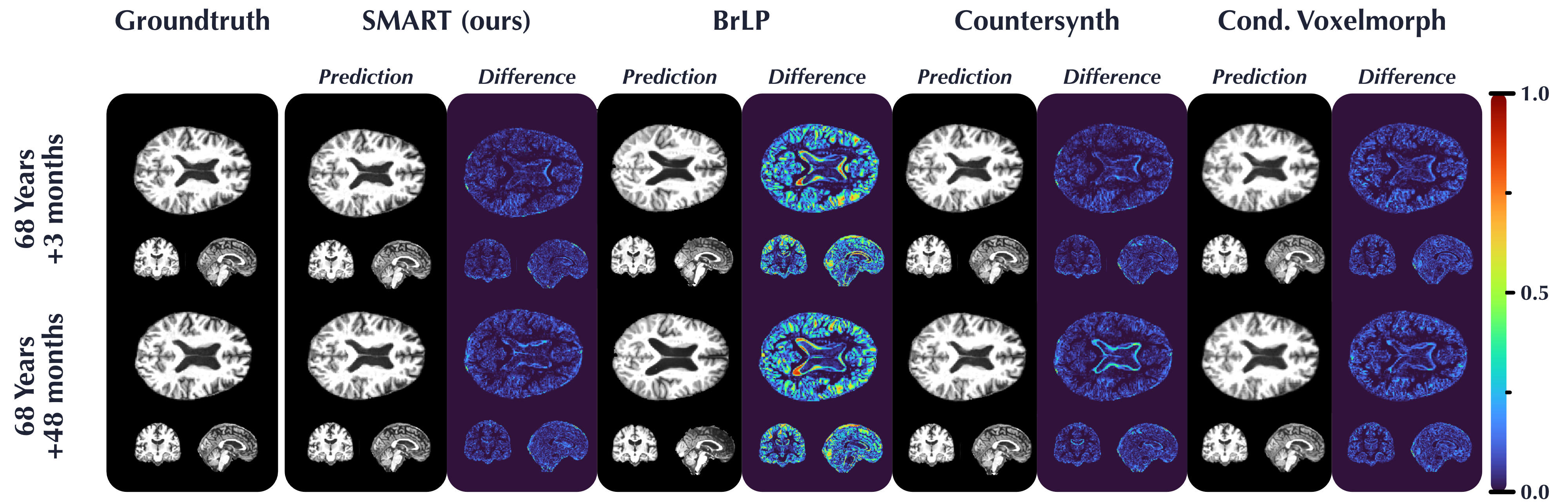}

   \caption{Qualitative comparison of the absolute error of \MNAME{} compared to the baselines BrLP, CounterSynth and Conditional Voxelmorph.}
   \label{fig:qualitative}
\end{figure*}

\subsection{Preprocessing - Turboprep}

We use the Turboprep pipeline \cite{turboprep2025} that consists of six steps to ensure consistent preprocessing across samples. It starts with intensity homogeneity normalisation \emph{N4} \cite{tustison2010n4itk}, Skull Stripping \emph{SynthStrip} \cite{hoopes2022synthstrip}, Affine registration \emph{SyN} \cite{avants2008symmetric} to template MNI152 \cite{HCPpipelines_MNI152_brain}, Segmentation of Brain Tissue \emph{SynthSeg 2.0} \cite{billot2023synthseg}, Brain Mask Extraction and lastly, Intensity normalisation with \emph{White Stripe} \cite{shinohara2014statistical}.

\subsection{Quantitative Results}

Table~\ref{tab:mse} shows that \MNAME{} achieves the best image-level accuracy (MSE $0.009$, SSIM $0.858$) despite using only 334K parameters. Representing change as a deformation rather than synthesizing intensities yields sharper anatomy and more stable temporal evolution. Region-wise MAE scores (Table~\ref{tab:mae}) highlight the main advantage: \MNAME{} captures anatomically relevant changes in most regions. On the contrary, CounterSynth consistently underestimates progression, while BrLP exhibits substantially higher errors across almost all regions. The conditional VoxelMorph baseline improves over diffusion approaches in several regions, but remains less accurate than \MNAME{}, particularly in regions with large structural changes such as white matter and ventricles. This trend is further supported by the supplementary shape-based evaluation, where \MNAME{} achieves the most consistent performance across Dice, HD95, and Fréchet distance, with the best results in 14/18, 17/18, and 14/18 regions, respectively.

\subsection{Qualitative Comparison}
Figure \ref{fig:qualitative} illustrates qualitative results comparing \MNAME{} with BrLP and CounterSynth across multiple timepoints. \MNAME{} produces anatomically coherent deformations that accurately follow the temporal progression of Alzheimer’s disease. Structural changes such as ventricular enlargement and hippocampal atrophy evolve smoothly and remain consistent with the predicted disease stage.

In contrast, CounterSynth tends to underestimate progression, generating follow-ups that remain overly close to the baseline appearance. BrLP, while producing plausible images, often fails to preserve subject-specific anatomy and instead converges toward an averaged, population-level representation.

\begin{table*}[htbp!]
\centering
\resizebox{\textwidth}{!}{

\begin{tabular}{lcccccc}
\hline
\textbf{Method} & Hip $\downarrow$ & Amyg $\downarrow$ & L.-Vent $\downarrow$ & C. WM $\downarrow$ & C. Cortex $\downarrow$ & Inf.-L. Vent $\downarrow$  \\
\hline
\MNAME{} (ours) & \textbf{0.012} $\pm$ \textbf{0.012} & \textbf{0.008} $\pm$ \textbf{0.007} & \textbf{0.176} $\pm$ \textbf{0.233}$^{*}$ & \textbf{0.399} $\pm$ \textbf{0.308}$^{*}$ & 0.433 $\pm$ 0.412 & \textbf{0.013} $\pm$ \textbf{0.024}$^{*}$  \\
No GPR & 0.015 $\pm$ 0.015 & 0.008 $\pm$ 0.007 & 0.240 $\pm$ 0.299 & 0.429 $\pm$ 0.330 & \textbf{0.428} $\pm$ \textbf{0.382} & 0.015 $\pm$ 0.025  \\
No TA & \textbf{0.012} $\pm$ \textbf{0.012} & \textbf{0.008} $\pm$ \textbf{0.007} & 0.182 $\pm$ 0.231 & 0.455 $\pm$ 0.326 & 0.480 $\pm$ 0.425 & 0.014 $\pm$ 0.023  \\
\hline
\\[-0.8em]
\textbf{Method} & Cereb. WM $\downarrow$ & Cereb. Cortex $\downarrow$ & Thalamus $\downarrow$ & Caudate $\downarrow$ & Putamen $\downarrow$ & Pallidum $\downarrow$ \\
\hline
\MNAME{} (ours) & \textbf{0.046} $\pm$ \textbf{0.040} & \textbf{0.097} $\pm$ \textbf{0.099}$^{*}$ & \textbf{0.015 $\pm$ 0.013}$^{*}$ & 0.020 $\pm$ 0.025 & \textbf{0.012} $\pm$ \textbf{0.010} & \textbf{0.007} $\pm$ \textbf{0.006}$^{*}$ \\
No GPR & \textbf{0.046} $\pm$ \textbf{0.039} & 0.102 $\pm$ 0.108 & 0.016 $\pm$ 0.014 & \textbf{0.011} $\pm$ \textbf{0.012}$^{*}$ & 0.013 $\pm$ 0.011 & 0.007 $\pm$ 0.006  \\
No TA & 0.053 $\pm$ 0.044 & 0.099 $\pm$ 0.100 & 0.017 $\pm$ 0.014 & 0.026 $\pm$ 0.027 & \textbf{0.012} $\pm$ \textbf{0.010} & 0.007 $\pm$ 0.006  \\
\hline
\\[-0.8em]
\textbf{Method} & Accumbens $\downarrow$ & VentralDC $\downarrow$ & Brain Stem $\downarrow$ & 3rd Vent $\downarrow$ & 4th Vent $\downarrow$ & CSF $\downarrow$ \\
\hline
\MNAME{} (ours) & \textbf{0.003} $\pm$ \textbf{0.003}$^{*}$ & \textbf{0.010} $\pm$ \textbf{0.008}$^{*}$ & \textbf{0.018} $\pm$ \textbf{0.018} & \textbf{0.007} $\pm$ \textbf{0.007}$^{*}$ &\textbf{0.005} $\pm$ \textbf{0.007} & 0.468 $\pm$ 0.420 \\
No GPR & 0.003 $\pm$ 0.003 & 0.011 $\pm$ 0.009 & \textbf{0.018} $\pm$ \textbf{0.018} & 0.008 $\pm$ 0.009 & 0.006 $\pm$ 0.007 & 0.519 $\pm$ 0.473  \\
No TA & 0.003 $\pm$ 0.003 & 0.010 $\pm$ 0.008 & \textbf{0.018} $\pm$ \textbf{0.018} & 0.007 $\pm$ 0.007 & \textbf{0.005} $\pm$ \textbf{0.007} & \textbf{0.444} $\pm$ \textbf{0.395}$^{*}$  \\
\hline
\end{tabular}
}
\caption{Ablation study of \MNAME{}, evaluating the impact of removing the temporal alignment (No TA) and removing the global progression model (No GPR). Metrics are reported as mean absolute error (MAE) of regional volumes expressed as a percentage of total brain volume (mean $\pm$ standard deviation); $*$ $p<0.05$ (pairwise test).}
\label{tab:mae_ablation}

\end{table*}

\subsection{Ablation Study}

We assess the contribution of the global progression model (GPR) and temporal alignment (TA). As shown in Table~\ref{tab:mae_ablation}, the full \MNAME{} achieves the lowest MAE, particularly in clinically relevant regions (e.g., hippocampus and ventricles). Removing GPR causes the largest degradation, highlighting the importance of region-wise progression modeling, while removing TA leads to smaller but consistent drops due to temporal misalignment. Differences in non-affected regions are negligible.

\begin{figure}[htbp]
  \centering
  \includegraphics[width=0.85\linewidth]{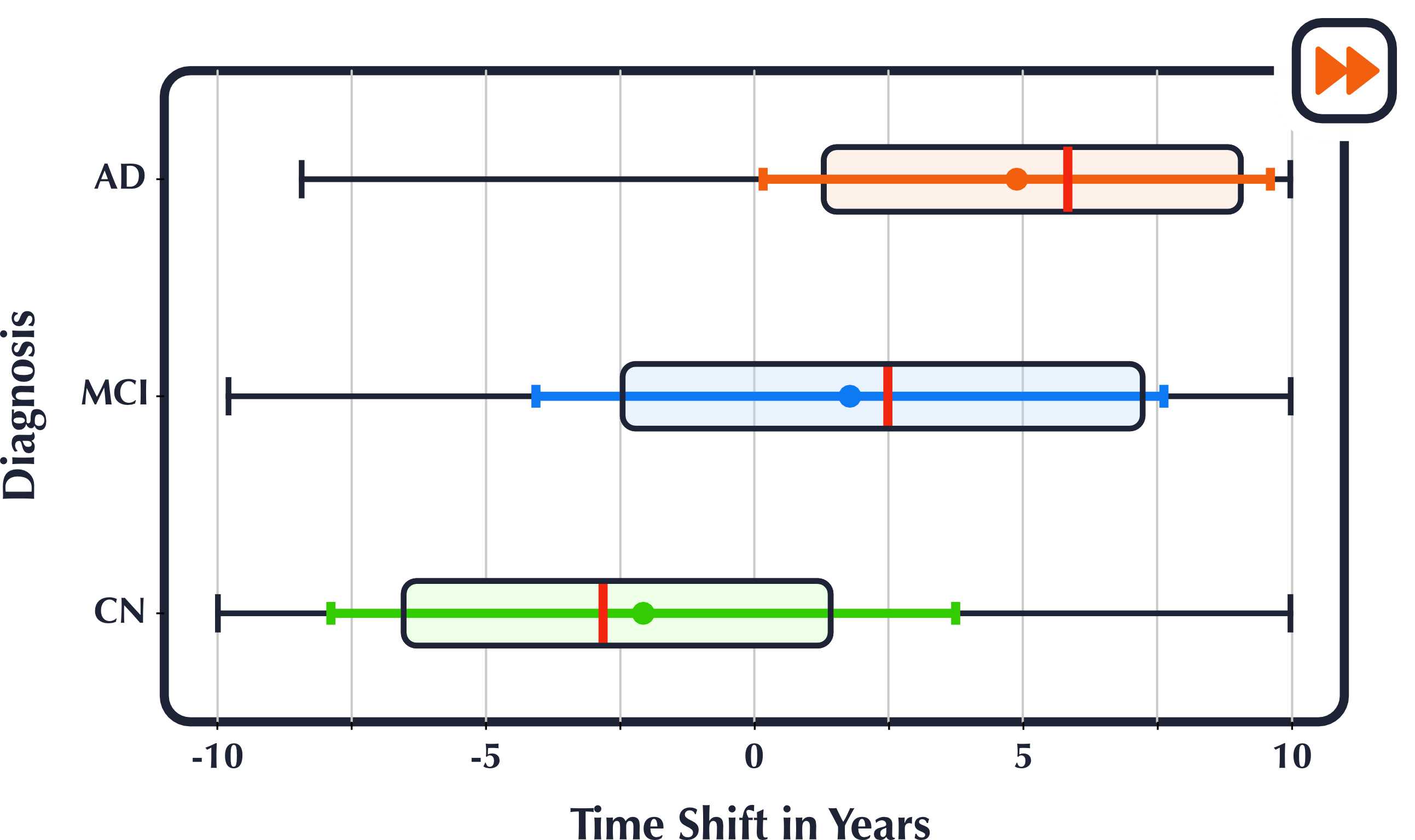}

   \caption{Distribution of predicted temporal alignments grouped by diagnostic category.}
   \label{fig:timeshift}
\end{figure}
\subsection{Analysis of the Individual Temporal Alignment}

To assess the relevance of the estimated {temporal alignments} across subjects, we examine their correlation with key demographic and clinical factors. Because detailed clinical measures are available only in the ADNI cohorts, correlations are computed exclusively on this subset using all available longitudinal time-points from the held-out subjects. The temporal alignments show a \emph{positive correlation with age} (\emph{Spearman} $rho = 0.46, p = 5.8 \times 10^{-20}$), indicating that the inferred disease stage increases with age. A similar positive correlation is observed with the {ADAS-13 cognitive score} ($rho = 0.32, p = 3.1 \times 10^{-8}$), suggesting that higher time shifts correspond to greater cognitive impairment. Conversely, the estimator exhibits a \emph{negative correlation with the $A\beta$ (amyloid-beta) in CSF level} ($rho = -0.41,\ p = 3.9 \times 10^{-4}$), consistent with expected patterns of neurodegeneration. In line with these results, a box-plot analysis (Figure \ref{fig:timeshift}) of the predicted temporal alignments across diagnostic groups (controls: CN, mild cognitive impairment: MCI, Alzheimer's disease: AD) shows a clear ordering of estimated disease stage, with progressively higher time shifts for more advanced diagnoses. These results suggest that the temporal alignment captures a meaningful representation of disease progression that aligns with both demographic and clinical markers across the entire course of Alzheimer’s pathology.

\begin{figure*}[htbp]
  \centering
  \includegraphics[width=0.90\linewidth]{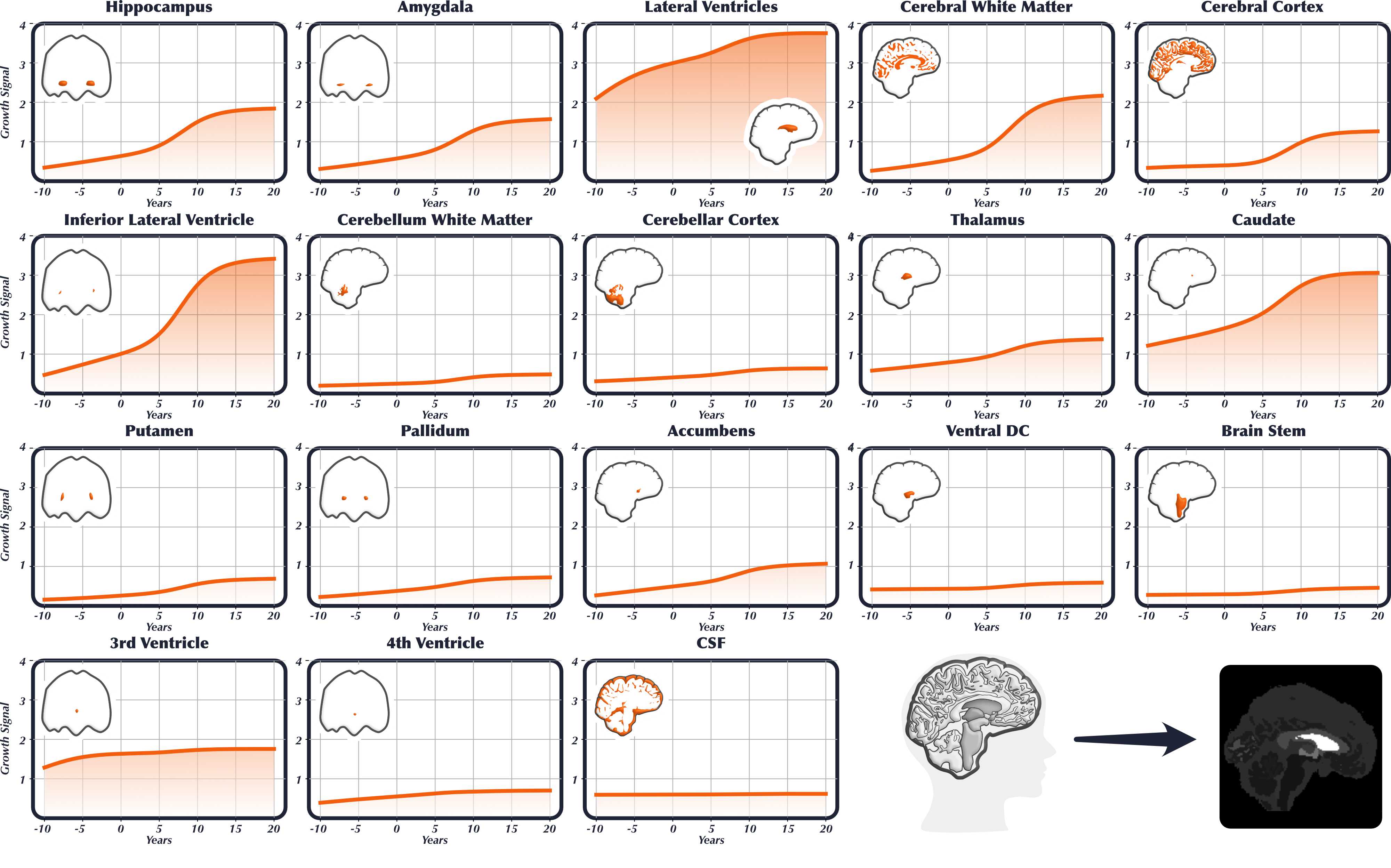}

   \caption{Region-wise progression functions learned by SMART. Each curve represents the estimated temporal trajectory of structural change for a specific brain region, parameterized as a continuous function of disease time. The functions capture distinct rates of atrophy or expansion, reflecting the heterogeneous progression dynamics of Alzheimer’s disease across anatomical regions.}
   \label{fig:funcProg}
\end{figure*}

\subsection{Learned Functional Progression}

We analyze the long-term regional disease progression curves learned by the spatio-temporal atlas, visualized in Figure \ref{fig:funcProg}. In spite of the short follow-up time across subjects (average 41 months), \MNAME{} identifies global group-wise progression patterns spanning more than 20 years. The dynamics of progression across regions reveal biologically consistent patterns of change: regions known to be strongly affected by Alzheimer’s disease, such as the hippocampus, lateral ventricles, inferior lateral ventricles, and third ventricle, exhibit pronounced trajectories of atrophy or expansion over time. In contrast, regions with limited atrophy in disease progression, such as cerebellum or brain stem, remain largely stable with low growth signals.

\section{Discussion}
\label{sec:discussion}

\MNAME{} demonstrates that disentangling temporal disease dynamics from spatial manifestation enables interpretable and efficient modeling of neurodegenerative progression. By formulating brain forecasting as a deformation process conditioned on region-wise ODE trajectories, the model combines statistical progression modeling and diffeomorphic registration, offering a unified view of how anatomical change unfolds over time. The resulting spatio-temporal atlas captures biologically plausible evolution patterns while remaining computationally lightweight and scalable to full-resolution volumes. 
A current limitation is that the learned progression functions represent \emph{cohort-averaged trajectories}, which do not yet account for \emph{disease subtypes}. Extending the framework to learn personalized or subgroup-specific dynamics, potentially through hierarchical modeling, represents a natural next step. Future work should evaluate the framework across other neurological disorders and broader disease settings, as well as over longer forecasting horizons. In addition, its behavior under treatment-induced deviations from natural disease progression remains to be explored, particularly in clinical trial scenarios where interventions may alter expected trajectories. 
Nevertheless, the present formulation provides a solid foundation for transparent, controllable disease modeling and marks a shift toward mechanistic representations of brain change in Alzheimer’s progression forecasting.

\section{Conclusion}
\label{sec:conclusion}

\textbf{\MNAME{}} reframes disease progression modeling as a structured \emph{spatio-temporal atlas estimation} problem rather than direct image synthesis.
By coupling continuous-time ODE dynamics with voxel-level Neural Cellular Automata, it learns a continuous disease-time atlas whose outputs are applied as anatomically coherent deformation fields capturing, region-specific progression patterns.
Across ADNI, OASIS-3, and AIBL, \MNAME{} surpasses adversarial and diffusion-based baselines in forecasting accuracy and temporal consistency.
The learned atlas recovers hallmark Alzheimer’s patterns, hippocampal shrinkage, ventricular expansion, and cortical thinning, and supports simulations of individual disease trajectories. 
By disentangling \emph{what changes, where, and when} from \emph{how those changes appear}, \MNAME{} establishes a flexible, interpretable, and scalable paradigm for modeling neurodegenerative processes through explicit disease-time atlas representations, moving beyond image-based prediction toward a more mechanistic understanding of brain evolution.

\section*{Acknowledgment}
\label{sec:acknowledgments}
This work was supported by the grants ANR-22-FAI1-0003 (TRAIN) and ANR-24-IAS2-0001 (Fed-Ops), by the PEPR Santé Numérique program through the grant ANR-22-PESN-0016 (REWIND), and by the 3IA Côte d’Azur initiatives, including ANR-19-P3IA-0002 (3IA Côte d’Azur) and ANR-23-IACL-0001 (3IA Côte d’Azur 2030).

\textbf{ADNI \cite{jack2008alzheimer}:} Data collection and sharing for the Alzheimer's Disease Neuroimaging Initiative (ADNI) is funded by the National Institute on Aging (National Institutes of Health Grant U19AG024904). The grantee organization is the Northern California Institute for Research and Education. In the past, ADNI has also received funding from the National Institute of Biomedical Imaging and Bioengineering, the Canadian Institutes of Health Research, and private sector contributions through the Foundation for the National Institutes of Health (FNIH) including generous contributions from the following: AbbVie, Alzheimer's Association; Alzheimer's Drug Discovery Foundation; Araclon Biotech; BioClinica, Inc.; Biogen; Bristol-Myers Squibb Company; CereSpir, Inc.; Cogstate; Eisai Inc.; Elan Pharmaceuticals, Inc.; Eli Lilly and Company; EuroImmun; F. Hoffmann-La Roche Ltd and its affiliated company Genentech, Inc.; Fujirebio; GE Healthcare; IXICO Ltd.; Janssen Alzheimer Immunotherapy Research \& Development, LLC.; Johnson \& Johnson Pharmaceutical Research \& Development LLC.; Lumosity; Lundbeck; Merck \& Co., Inc.; Meso Scale Diagnostics, LLC.; NeuroRx Research; Neurotrack Technologies; Novartis Pharmaceuticals Corporation; Pfizer Inc.; Piramal Imaging; Servier; Takeda Pharmaceutical Company; and Transition Therapeutics. 

\textbf{OASIS \cite{lamontagne2019oasis}:} Data were provided in part by OASIS-3: Longitudinal Multimodal Neuroimaging: Principal Investigators: T. Benzinger, D. Marcus, J. Morris; NIH P30 AG066444, P50 AG00561, P30 NS09857781, P01 AG026276, P01 AG003991, R01 AG043434, UL1 TR000448, R01 EB009352.

{
    \small
    \bibliographystyle{ieeenat_fullname}
    \bibliography{main}

\begin{thebibliography}{42}
\providecommand{\natexlab}[1]{#1}
\providecommand{\url}[1]{\texttt{#1}}
\expandafter\ifx\csname urlstyle\endcsname\relax
  \providecommand{\doi}[1]{doi: #1}\else
  \providecommand{\doi}{doi: \begingroup \urlstyle{rm}\Url}\fi

\bibitem[Abi~Nader et~al.(2021)Abi~Nader, Ayache, Frisoni, Robert, Lorenzi, and
  {A}lzheimer’s Disease Neuroimaging~Initiative]{abi2021simulating}
Cl{\'e}ment Abi~Nader, Nicholas Ayache, Giovanni~B Frisoni, Philippe Robert,
  Marco Lorenzi, and {A}lzheimer’s Disease Neuroimaging~Initiative.
\newblock Simulating the outcome of amyloid treatments in {A}lzheimer's disease
  from imaging and clinical data.
\newblock \emph{Brain communications}, 3\penalty0 (2):\penalty0 fcab091, 2021.

\bibitem[Avants et~al.(2008)Avants, Epstein, Grossman, and
  Gee]{avants2008symmetric}
Brian~B Avants, Charles~L Epstein, Murray Grossman, and James~C Gee.
\newblock Symmetric diffeomorphic image registration with cross-correlation:
  evaluating automated labeling of elderly and neurodegenerative brain.
\newblock \emph{Medical image analysis}, 12\penalty0 (1):\penalty0 26--41,
  2008.

\bibitem[Balakrishnan et~al.(2019)Balakrishnan, Zhao, Sabuncu, Guttag, and
  Dalca]{balakrishnan2019voxelmorph}
Guha Balakrishnan, Amy Zhao, Mert~R Sabuncu, John Guttag, and Adrian~V Dalca.
\newblock Voxelmorph: a learning framework for deformable medical image
  registration.
\newblock \emph{IEEE transactions on medical imaging}, 38\penalty0
  (8):\penalty0 1788--1800, 2019.

\bibitem[Billot et~al.(2023)Billot, Greve, Puonti, Thielscher, Van~Leemput,
  Fischl, Dalca, Iglesias, et~al.]{billot2023synthseg}
Benjamin Billot, Douglas~N Greve, Oula Puonti, Axel Thielscher, Koen
  Van~Leemput, Bruce Fischl, Adrian~V Dalca, Juan~Eugenio Iglesias, et~al.
\newblock {SynthSeg}: Segmentation of brain mri scans of any contrast and
  resolution without retraining.
\newblock \emph{Medical image analysis}, 86:\penalty0 102789, 2023.

\bibitem[Chen et~al.(2018)Chen, Rubanova, Bettencourt, and
  Duvenaud]{chen2018neural}
Ricky~TQ Chen, Yulia Rubanova, Jesse Bettencourt, and David~K Duvenaud.
\newblock Neural ordinary differential equations.
\newblock \emph{Advances in neural information processing systems}, 31, 2018.

\bibitem[Cui et~al.(2019)Cui, Liu, {A}lzheimer's Disease
  Neuroimaging~Initiative, et~al.]{cui2019rnn}
Ruoxuan Cui, Manhua Liu, {A}lzheimer's Disease Neuroimaging~Initiative, et~al.
\newblock {RNN}-based longitudinal analysis for diagnosis of {A}lzheimer’s
  disease.
\newblock \emph{Computerized Medical Imaging and Graphics}, 73:\penalty0 1--10,
  2019.

\bibitem[Dalca et~al.(2019)Dalca, Rakic, Guttag, and
  Sabuncu]{dalca2019learning}
Adrian Dalca, Marianne Rakic, John Guttag, and Mert Sabuncu.
\newblock Learning conditional deformable templates with convolutional
  networks.
\newblock \emph{Advances in neural information processing systems}, 32, 2019.

\bibitem[Donohue et~al.(2014)Donohue, Jacqmin-Gadda, Le~Goff, Thomas, Raman,
  Gamst, Beckett, Jack~Jr, Weiner, Dartigues, et~al.]{donohue2014estimating}
Michael~C Donohue, H{\'e}l{\`e}ne Jacqmin-Gadda, M{\'e}lanie Le~Goff, Ronald~G
  Thomas, Rema Raman, Anthony~C Gamst, Laurel~A Beckett, Clifford~R Jack~Jr,
  Michael~W Weiner, Jean-Fran{\c{c}}ois Dartigues, et~al.
\newblock Estimating long-term multivariate progression from short-term data.
\newblock \emph{{A}lzheimer's \& Dementia}, 10:\penalty0 S400--S410, 2014.

\bibitem[Durrleman et~al.(2009)Durrleman, Pennec, Trouv{\'e}, Gerig, and
  Ayache]{durrleman2009spatiotemporal}
Stanley Durrleman, Xavier Pennec, Alain Trouv{\'e}, Guido Gerig, and Nicholas
  Ayache.
\newblock Spatiotemporal atlas estimation for developmental delay detection in
  longitudinal datasets.
\newblock In \emph{International Conference on Medical Image Computing and
  Computer-Assisted Intervention}, pages 297--304. Springer, 2009.

\bibitem[Ellis et~al.(2009)Ellis, Bush, Darby, De~Fazio, Foster, Hudson,
  Lautenschlager, Lenzo, Martins, Maruff, et~al.]{ellis2009australian}
Kathryn~A Ellis, Ashley~I Bush, David Darby, Daniela De~Fazio, Jonathan Foster,
  Peter Hudson, Nicola~T Lautenschlager, Nat Lenzo, Ralph~N Martins, Paul
  Maruff, et~al.
\newblock The australian imaging, biomarkers and lifestyle (aibl) study of
  aging: methodology and baseline characteristics of 1112 individuals recruited
  for a longitudinal study of {A}lzheimer's disease.
\newblock \emph{International psychogeriatrics}, 21\penalty0 (4):\penalty0
  672--687, 2009.

\bibitem[Fox and Schott(2004)]{fox2004imaging}
Nick~C Fox and Jonathan~M Schott.
\newblock Imaging cerebral atrophy: normal ageing to {A}lzheimer's disease.
\newblock \emph{The Lancet}, 363\penalty0 (9406):\penalty0 392--394, 2004.

\bibitem[Hoopes et~al.(2022)Hoopes, Mora, Dalca, Fischl, and
  Hoffmann]{hoopes2022synthstrip}
Andrew Hoopes, Jocelyn~S Mora, Adrian~V Dalca, Bruce Fischl, and Malte
  Hoffmann.
\newblock {SynthStrip}: skull-stripping for any brain image.
\newblock \emph{NeuroImage}, 260:\penalty0 119474, 2022.

\bibitem[Jack~Jr et~al.(2008)Jack~Jr, Bernstein, Fox, Thompson, Alexander,
  Harvey, Borowski, Britson, L.~Whitwell, Ward, et~al.]{jack2008alzheimer}
Clifford~R Jack~Jr, Matt~A Bernstein, Nick~C Fox, Paul Thompson, Gene
  Alexander, Danielle Harvey, Bret Borowski, Paula~J Britson, Jennifer
  L.~Whitwell, Chadwick Ward, et~al.
\newblock The {A}lzheimer's disease neuroimaging initiative (adni): Mri
  methods.
\newblock \emph{Journal of Magnetic Resonance Imaging: An Official Journal of
  the International Society for Magnetic Resonance in Medicine}, 27\penalty0
  (4):\penalty0 685--691, 2008.

\bibitem[Jian et~al.(2026)Jian, Pan, Li, Bongratz, Li, Rueckert, Wiestler, and
  Wachinger]{jian2026temporal}
Bailiang Jian, Jiazhen Pan, Yitong Li, Fabian Bongratz, Ruochen Li, Daniel
  Rueckert, Benedikt Wiestler, and Christian Wachinger.
\newblock Temporal conditioning for longitudinal brain mri registration and
  aging analysis.
\newblock \emph{IEEE Transactions on Medical Imaging}, 2026.

\bibitem[Kalkhof and Mukhopadhyay(2023)]{kalkhof2023m3d}
John Kalkhof and Anirban Mukhopadhyay.
\newblock {M3D-NCA}: Robust 3d segmentation with built-in quality control.
\newblock In \emph{International Conference on Medical Image Computing and
  Computer-Assisted Intervention}, pages 169--178. Springer, 2023.

\bibitem[Kalkhof et~al.(2023)Kalkhof, Gonz{\'a}lez, and
  Mukhopadhyay]{kalkhof2023med}
John Kalkhof, Camila Gonz{\'a}lez, and Anirban Mukhopadhyay.
\newblock {Med-NCA}: Robust and lightweight segmentation with neural cellular
  automata.
\newblock In \emph{International Conference on Information Processing in
  Medical Imaging}, pages 705--716. Springer, 2023.

\bibitem[Kalkhof et~al.(2025)Kalkhof, Ihm, K{\"o}hler, Gregori, and
  Mukhopadhyay]{kalkhof2025med}
John Kalkhof, Niklas Ihm, Tim K{\"o}hler, Bjarne Gregori, and Anirban
  Mukhopadhyay.
\newblock {MED-NCA}: Bio-inspired medical image segmentation.
\newblock \emph{Medical Image Analysis}, page 103601, 2025.

\bibitem[Koval et~al.(2021)Koval, B{\^o}ne, Louis, Lartigue, Bottani, Marcoux,
  Samper-Gonzalez, Burgos, Charlier, Bertrand, et~al.]{koval2021ad}
Igor Koval, Alexandre B{\^o}ne, Maxime Louis, Thomas Lartigue, Simona Bottani,
  Arnaud Marcoux, Jorge Samper-Gonzalez, Ninon Burgos, Benjamin Charlier, Anne
  Bertrand, et~al.
\newblock {AD} course map charts {A}lzheimer’s disease progression.
\newblock \emph{Scientific Reports}, 11\penalty0 (1):\penalty0 8020, 2021.

\bibitem[Lachinov et~al.(2023)Lachinov, Chakravarty, Grechenig,
  Schmidt-Erfurth, and Bogunovi{\'c}]{lachinov2023learning}
Dmitrii Lachinov, Arunava Chakravarty, Christoph Grechenig, Ursula
  Schmidt-Erfurth, and Hrvoje Bogunovi{\'c}.
\newblock Learning spatio-temporal model of disease progression with neuralodes
  from longitudinal volumetric data.
\newblock \emph{IEEE Transactions on Medical Imaging}, 43\penalty0
  (3):\penalty0 1165--1179, 2023.

\bibitem[LaMontagne et~al.(2019)LaMontagne, Benzinger, Morris, Keefe, Hornbeck,
  Xiong, Grant, Hassenstab, Moulder, Vlassenko, et~al.]{lamontagne2019oasis}
Pamela~J LaMontagne, Tammie~LS Benzinger, John~C Morris, Sarah Keefe, Russ
  Hornbeck, Chengjie Xiong, Elizabeth Grant, Jason Hassenstab, Krista Moulder,
  Andrei~G Vlassenko, et~al.
\newblock Oasis-3: longitudinal neuroimaging, clinical, and cognitive dataset
  for normal aging and {A}lzheimer disease.
\newblock \emph{medrxiv}, pages 2019--12, 2019.

\bibitem[Lorenzi et~al.(2011)Lorenzi, Ayache, Frisoni, Pennec, and
  ADNI]{lorenzi2011mapping}
Marco Lorenzi, Nicholas Ayache, Giovanni~B Frisoni, Xavier Pennec, and ADNI.
\newblock Mapping the effects of a $\beta$ 1- 42 levels on the longitudinal
  changes in healthy aging: Hierarchical modeling based on stationary velocity
  fields.
\newblock In \emph{International Conference on Medical Image Computing and
  Computer-Assisted Intervention}, pages 663--670. Springer, 2011.

\bibitem[Lorenzi et~al.(2019)Lorenzi, Filippone, Frisoni, Alexander, Ourselin,
  {A}lzheimer's Disease Neuroimaging~Initiative,
  et~al.]{lorenzi2019probabilistic}
Marco Lorenzi, Maurizio Filippone, Giovanni~B Frisoni, Daniel~C Alexander,
  S{\'e}bastien Ourselin, {A}lzheimer's Disease Neuroimaging~Initiative, et~al.
\newblock Probabilistic disease progression modeling to characterize diagnostic
  uncertainty: application to staging and prediction in {A}lzheimer's disease.
\newblock \emph{NeuroImage}, 190:\penalty0 56--68, 2019.

\bibitem[Marti-Juan et~al.(2023)Marti-Juan, Lorenzi, Piella, {A}lzheimer’s
  Disease Neuroimaging~Initiative, et~al.]{marti2023mc}
Gerard Marti-Juan, Marco Lorenzi, Gemma Piella, {A}lzheimer’s Disease
  Neuroimaging~Initiative, et~al.
\newblock {MC-RVAE}: multi-channel recurrent variational autoencoder for
  multimodal {A}lzheimer’s disease progression modelling.
\newblock \emph{NeuroImage}, 268:\penalty0 119892, 2023.

\bibitem[Miller et~al.(2002)Miller, Trouv{\'e}, and Younes]{miller2002metrics}
Michael~I Miller, Alain Trouv{\'e}, and Laurent Younes.
\newblock On the metrics and euler-lagrange equations of computational anatomy.
\newblock \emph{Annual review of biomedical engineering}, 4\penalty0
  (1):\penalty0 375--405, 2002.

\bibitem[Mordvintsev et~al.(2020)Mordvintsev, Randazzo, Niklasson, and
  Levin]{mordvintsev2020growing}
Alexander Mordvintsev, Ettore Randazzo, Eyvind Niklasson, and Michael Levin.
\newblock Growing neural cellular automata.
\newblock \emph{Distill}, 5\penalty0 (2):\penalty0 e23, 2020.

\bibitem[Niethammer et~al.(2011)Niethammer, Huang, and
  Vialard]{niethammer2011geodesic}
Marc Niethammer, Yang Huang, and Fran{\c{c}}ois-Xavier Vialard.
\newblock Geodesic regression for image time-series.
\newblock In \emph{International conference on medical image computing and
  computer-assisted intervention}, pages 655--662. Springer, 2011.

\bibitem[Petersen et~al.(2010)Petersen, Aisen, Beckett, Donohue, Gamst, Harvey,
  Jack~Jr, Jagust, Shaw, Toga, et~al.]{petersen2010alzheimer}
Ronald~Carl Petersen, Paul~S Aisen, Laurel~A Beckett, Michael~C Donohue,
  Anthony~Collins Gamst, Danielle~J Harvey, CR Jack~Jr, William~J Jagust,
  Leslie~M Shaw, Arthur~W Toga, et~al.
\newblock {A}lzheimer's disease neuroimaging initiative (adni) clinical
  characterization.
\newblock \emph{Neurology}, 74\penalty0 (3):\penalty0 201--209, 2010.

\bibitem[Pombo et~al.(2023)Pombo, Gray, Cardoso, Ourselin, Rees, Ashburner, and
  Nachev]{pombo2023equitable}
Guilherme Pombo, Robert Gray, M~Jorge Cardoso, Sebastien Ourselin, Geraint
  Rees, John Ashburner, and Parashkev Nachev.
\newblock Equitable modelling of brain imaging by counterfactual augmentation
  with morphologically constrained 3d deep generative models.
\newblock \emph{Medical Image Analysis}, 84:\penalty0 102723, 2023.

\bibitem[Puglisi(2023)]{turboprep2025}
Lemuel Puglisi.
\newblock Turboprep.
\newblock \url{https://github.com/LemuelPuglisi/turboprep}, 2023.
\newblock Accessed: 2025-10-08.

\bibitem[Puglisi et~al.(2024)Puglisi, Alexander, and
  Rav{\`\i}]{puglisi2024enhancing}
Lemuel Puglisi, Daniel~C Alexander, and Daniele Rav{\`\i}.
\newblock Enhancing spatiotemporal disease progression models via latent
  diffusion and prior knowledge.
\newblock In \emph{International Conference on Medical Image Computing and
  Computer-Assisted Intervention}, pages 173--183. Springer, 2024.

\bibitem[Qiu et~al.(2009)Qiu, Albert, Younes, and Miller]{qiu2009time}
Anqi Qiu, Marilyn Albert, Laurent Younes, and Michael~I Miller.
\newblock Time sequence diffeomorphic metric mapping and parallel transport
  track time-dependent shape changes.
\newblock \emph{NeuroImage}, 45\penalty0 (1):\penalty0 S51--S60, 2009.

\bibitem[Ranem et~al.(2024)Ranem, Kalkhof, and Mukhopadhyay]{ranem2024nca}
Amin Ranem, John Kalkhof, and Anirban Mukhopadhyay.
\newblock {NCA-Morph}: Medical image registration with neural cellular
  automata.
\newblock \emph{arXiv preprint arXiv:2410.22265}, 2024.

\bibitem[Ravi et~al.(2022)Ravi, Blumberg, Ingala, Barkhof, Alexander, Oxtoby,
  {A}lzheimer’s Disease Neuroimaging~Initiative,
  et~al.]{ravi2022degenerative}
Daniele Ravi, Stefano~B Blumberg, Silvia Ingala, Frederik Barkhof, Daniel~C
  Alexander, Neil~P Oxtoby, {A}lzheimer’s Disease Neuroimaging~Initiative,
  et~al.
\newblock Degenerative adversarial neuroimage nets for brain scan simulations:
  Application in ageing and dementia.
\newblock \emph{Medical Image Analysis}, 75:\penalty0 102257, 2022.

\bibitem[Schuster and Paliwal(1997)]{schuster1997bidirectional}
Mike Schuster and Kuldip~K Paliwal.
\newblock Bidirectional recurrent neural networks.
\newblock \emph{IEEE transactions on Signal Processing}, 45\penalty0
  (11):\penalty0 2673--2681, 1997.

\bibitem[Shaw et~al.(2009)Shaw, Vanderstichele, Knapik-Czajka, Clark, Aisen,
  Petersen, Blennow, Soares, Simon, Lewczuk, et~al.]{shaw2009cerebrospinal}
Leslie~M Shaw, Hugo Vanderstichele, Malgorzata Knapik-Czajka, Christopher~M
  Clark, Paul~S Aisen, Ronald~C Petersen, Kaj Blennow, Holly Soares, Adam
  Simon, Piotr Lewczuk, et~al.
\newblock Cerebrospinal fluid biomarker signature in {A}lzheimer's disease
  neuroimaging initiative subjects.
\newblock \emph{Annals of neurology}, 65\penalty0 (4):\penalty0 403--413, 2009.

\bibitem[Shinohara et~al.(2014)Shinohara, Sweeney, Goldsmith, Shiee, Mateen,
  Calabresi, Jarso, Pham, Reich, Crainiceanu, et~al.]{shinohara2014statistical}
Russell~T Shinohara, Elizabeth~M Sweeney, Jeff Goldsmith, Navid Shiee, Farrah~J
  Mateen, Peter~A Calabresi, Samson Jarso, Dzung~L Pham, Daniel~S Reich,
  Ciprian~M Crainiceanu, et~al.
\newblock Statistical normalization techniques for magnetic resonance imaging.
\newblock \emph{NeuroImage: Clinical}, 6:\penalty0 9--19, 2014.

\bibitem[Thompson et~al.(2007)Thompson, Hayashi, Dutton, CHIANG, Leow, Sowell,
  De~Zubicaray, Becker, Lopez, Aizenstein, et~al.]{thompson2007tracking}
Paul~M Thompson, Kiralee~M Hayashi, Rebecca~A Dutton, MING-CHANG CHIANG, Alex~D
  Leow, Elizabeth~R Sowell, Greig De~Zubicaray, James~T Becker, Oscar~L Lopez,
  Howard~J Aizenstein, et~al.
\newblock Tracking {A}lzheimer's disease.
\newblock \emph{Annals of the New York Academy of Sciences}, 1097\penalty0
  (1):\penalty0 183--214, 2007.

\bibitem[Tustison et~al.(2010)Tustison, Avants, Cook, Zheng, Egan, Yushkevich,
  and Gee]{tustison2010n4itk}
Nicholas~J Tustison, Brian~B Avants, Philip~A Cook, Yuanjie Zheng, Alexander
  Egan, Paul~A Yushkevich, and James~C Gee.
\newblock {N4ITK}: improved n3 bias correction.
\newblock \emph{IEEE transactions on medical imaging}, 29\penalty0
  (6):\penalty0 1310--1320, 2010.

\bibitem[Vercauteren et~al.(2007)Vercauteren, Pennec, Perchant, and
  Ayache]{vercauteren2007non}
Tom Vercauteren, Xavier Pennec, Aymeric Perchant, and Nicholas Ayache.
\newblock Non-parametric diffeomorphic image registration with the demons
  algorithm.
\newblock In \emph{International conference on medical image computing and
  computer-assisted intervention}, pages 319--326. Springer, 2007.

\bibitem[Von~Neumann et~al.(1966)Von~Neumann, Burks, et~al.]{von1966theory}
John Von~Neumann, Arthur~Walter Burks, et~al.
\newblock Theory of self-reproducing automata.
\newblock 1966.

\bibitem[{Washington University HCPpipelines}(2025)]{HCPpipelines_MNI152_brain}
{Washington University HCPpipelines}.
\newblock {MNI152}\_{T1}\_1mm\_brain.nii.gz.
\newblock GitHub repository,
  \url{https://github.com/Washington-University/HCPpipelines}, 2025.
\newblock Template file in global/templates; Accessed: 2025-10-08.

\bibitem[Yoon et~al.(2023)Yoon, Zhang, Suk, Guo, and Li]{yoon2023sadm}
Jee~Seok Yoon, Chenghao Zhang, Heung-Il Suk, Jia Guo, and Xiaoxiao Li.
\newblock {SADM}: Sequence-aware diffusion model for longitudinal medical image
  generation.
\newblock In \emph{International Conference on Information Processing in
  Medical Imaging}, pages 388--400. Springer, 2023.

\end{thebibliography}
}


\end{document}